\documentclass[runningheads]{llncs}
\usepackage{amsmath}
\usepackage{multirow}
\usepackage{graphicx}
\begin{document}

\title{Multi-task Learning for Low-resource Second Language Acquisition Modeling}

\author{Yong Hu\inst{1,2} \and
Heyan Huang\inst{1} \and
Tian Lan\inst{1} \and
Xiaochi Wei\inst{3} \and
Yuxiang Nie\inst{1} \and
Jiarui Qi\inst{1} \and
Liner Yang\inst{4} \and
Xian-Ling Mao\inst{1}
}
\authorrunning{Yong Hu, Heyan Huang et al.}

\institute{
  School of Computer Science and Technology, Beijing Institute of Technology, China \and
  CETC Big DataResearch Institute Co., Ltd., Guiyang, China \and
  Baidu Inc., Beijing, China  \and
  Beijing Language and Culture University, Beijing, China
  \email{\{huyong,hhy63,maoxl\}@bit.edu.cn}, \email{weixiaochi@baidu.com} \\
  \email{\{lantiangmftby,jerrrynie,Rita2663269,lineryang\}@gmail.com}
}

\maketitle              
\begin{abstract}
Second language acquisition (SLA) modeling is to predict whether second language learners could correctly answer the questions according to what they have learned. 
It is a fundamental building block of the personalized learning system and has attracted more and more attention recently.  
However, as far as we know, almost all existing methods cannot work well in low-resource scenarios because lacking of training data.
Fortunately, there are some latent common patterns among different language-learning tasks, which gives us an opportunity to solve the low-resource SLA modeling problem.  
Inspired by this idea, in this paper, we propose a novel SLA modeling method, which learns the latent common patterns among different language-learning datasets by multi-task learning and are further applied to improving the prediction performance in low-resource scenarios.  
Extensive experiments show that the proposed method performs much better than the state-of-the-art baselines in the low-resource scenario. Meanwhile, it also obtains improvement slightly in the non-low-resource scenario.

\end{abstract}
\section{Introduction}
Knowledge tracing (KT) is a task of modeling how much knowledge students have obtained over time so that we can accurately predict how students will perform on future exercises and arrange study plans dynamically according to their real-time situations \cite{bauman2014recommending,pelanek2017bayesian}.
Particularly, second language acquisition (SLA) modeling is a kind of KT in the filed of language learning.
With the increasing importance of language-learning activity in people's daily life \cite{larsen2014introduction}, 
SLA modeling attracts more and more attention. For example, NAACL 2018 had held a public SLA modeling challenge.\footnote{http://sharedtask.duolingo.com/2018.html}  
Therefore, in this paper, we focus on SLA modeling.

\begin{figure}[t!]
  \centering
  \includegraphics[width=0.8\textwidth]{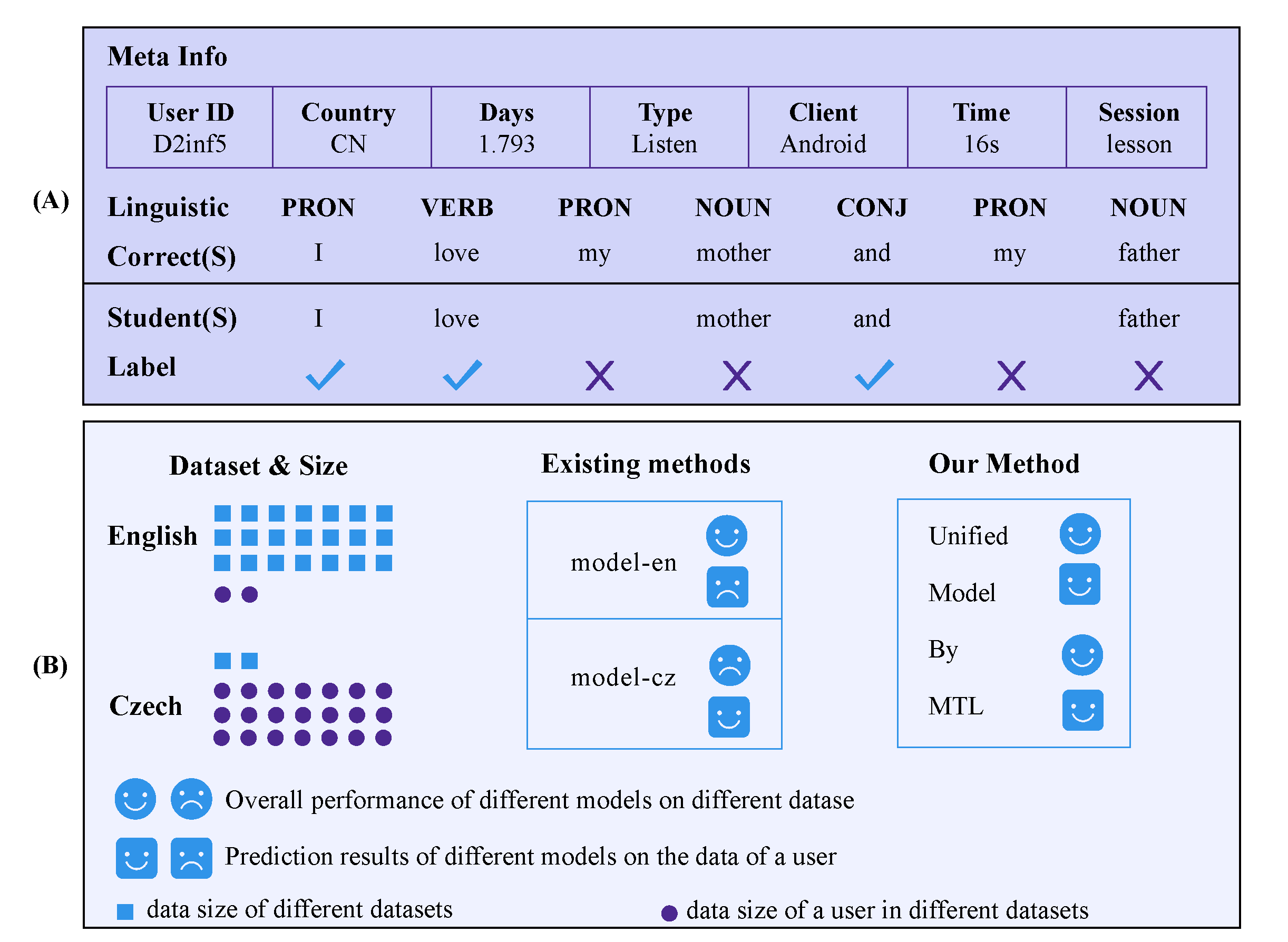}
\caption{(A) Illustration of an example of SLA modeling task. (B) Illustration of two kinds of low-resource phenomenons and the comparison of our method and existing methods.}
\label{Fig:idea}
\end{figure}

SLA modeling is the learning process of a specific language, thus each SLA modeling task has a corresponding language, e.g., English, Spanish, and French. Meanwhile, each language is composed of many exercises, and an exercise is the smallest data unit. 
For an exercise, there are three possible types, i.e., \textit{listen}, \textit{Translation}, and \textit{Reverse Tap}, and the answers to the exercises are all sentences regardless of the type of the exercise.
In an exercise, a student will answer the given question and write its answer sentence.
Then the student-provided sentence and the correct sentence will be compared word by word to evaluate the ability of the student.
As shown in Fig.~\ref{Fig:idea} (A), taking an English listening exercise as an example, the correct sentence is `` \textit{I love my mother and my father}", and the answer of the student is `` \textit{I love mader and fhader}"; It can be shown that there are three words that are correctly answered.
Therefore, SLA modeling task is to predict whether students can answer each word correctly according to the exercise information (meta-information, correct sentence with corresponding linguistic information). Thus, it can be simply token into a word-level binary classification task.

In SLA modeling task, low-resource is a common phenomenon which affects the training process significantly. Specifically, this phenomena is mainly caused by two reasons:
(1) For some specific language-learning datasets, e.g. Czech, the size of data may be very small because we cannot collect enough language-learning exercises;
(2) For a user, he/she will encounter cold start scenario when starting to learn a new language.
However, almost all existing methods for SLA modeling task train a model separately for each language-learning dataset and thus their performance largely depends on the size of training data. Thus, they can hardly work well in low-resource scenarios.
Fig.~\ref{Fig:idea} (B) illustrates an example.
Suppose that we have two language: English and Czech, existing methods will train two separate models for these two languages: \textit{model\_en} and \textit{model\_cz}. 
These two models will perform poorly in two low-resource scenarios:
(1) If the English dataset has a large amount of data, the \textit{model\_en} will perform well, but the small size of Czech dataset may significantly hinders the performance of \textit{model\_cz};
(2) Suppose that a user has a large number of exercises for learning Czech, but when he/she begins to learn English, the number of English exercises for him/her will be very small, even zero. Thus, \textit{model\_en} can hardly predict the answers of his/her English exercises well.

Intuitively, there are lots of common patterns among different language-learning tasks, such as the learning habits of users and grammar learning skills.  
If the latent common patterns across these language-learning tasks can be well learned, 
they can be used to solve the low-resource SLA modeling problem.

Inspired by this idea, in this paper, we propose a novel multi-task learning method for SLA modeling, which is a unified model to process several language-learning datasets simultaneously.
Specifically, the proposed model learns shared features across all language-learning datasets jointly, which is the inner nature of the language-learning activity, 
and can be taken as important prior-knowledge to deal with small language-learning datasets.
Moreover, the embedding information of a user is shared, so the learning habits and language talents of the user could be shared in the unified model for other low-resource language-learning tasks.  
Therefore, when a user begins to learn a new language, the unified model can work well even though there is no exercise data for this user. 

The main contributions of this paper are three-fold. (1) As
far as we know, this is the first work applying multi-task neural network to SLA modeling and we effectively solve the problem of insufficient training data in low-resource scenarios.
(2) We deeply study the common patterns among different languages and reveal the inner nature of language learning.
(3) Extensive experiments show that our method performs much better than the state-of-the-art baselines in low-resource scenarios, and it also obtains improvement slightly in the non-low-resource scenario. 
Additionally, we have publicly released our codes to facilitate follow-on researchers.\footnote{\url{https://github.com/nghuyong/MTL-SLAM}}

\section{Related Work}
\subsection{SLA Modeling}
Existing methods for SLA modeling can be roughly divided into three categories: (a) logistic regression based methods, (b) tree ensemble methods, and (c) sequence modeling methods.
(a) The logistic regression based methods \cite{klerke2018grotoco,nayak2018context} take the meta and context features provided by datasets and other manually constructed features as input and output the probability of answering each word correctly. These methods are simple but their performances are not very poor.
(b) The tree ensemble methods (e.g., Gradient Boosting Decision Trees (GBDT)) \cite{tomoschuk2018memory,rich2018modeling} can powerfully capture non-linear relationships between features.
Therefore, although the input and output of these methods are the same with (a), they are generally better than methods that belong to (a).
(c) The sequence modeling methods (e.g., Recurrent Neural Networks (RNNs))  \cite{xu2018cluf,yuan2018neural} use neural networks, especially RNNs so that they can capture users' performance over time. The performance of these methods are also very competitive.

However, methods above hardly can work well in low-resource scenarios because their performance largely depends on the size of training data.

\subsection{Multi-Task Learning}
Multi-task learning (MTL) has been widely used in various tasks, such as machine learning\cite{liu2019intelligent,he2018similarity,jiang2016novel}, natural language processing \cite{collobert2008unified,liu2016recurrent,dong2015multi}, speech recognition \cite{deng2013new,kim2017joint,wu2015deep} and computer vision \cite{chen2018multi,guo2015human,zhang2014facial}.
It effectively increases the sample size that we are using for training our model. Thus, it can improve generalization by leveraging the domain-specific information contained in related tasks, and enables the model to obtain a better sharing representation between each related task.

MTL is typically done with hard or soft parameter sharing of hidden layers and hard parameter sharing is the most commonly used approach to MTL in neural networks \cite{DBLP:journals/corr/Ruder17a}. It is generally applied by sharing the hidden layers between all tasks, while keeping several task-specific output layers.

SLA modeling has different language-learning tasks, and each task has something in common, which gives us an opportunity to use MTL to improve the overall performance.

\section{Model}
\subsection{Problem Definition}

\begin{figure*}[t!]
  \centering
  \includegraphics[width=1.0\textwidth]{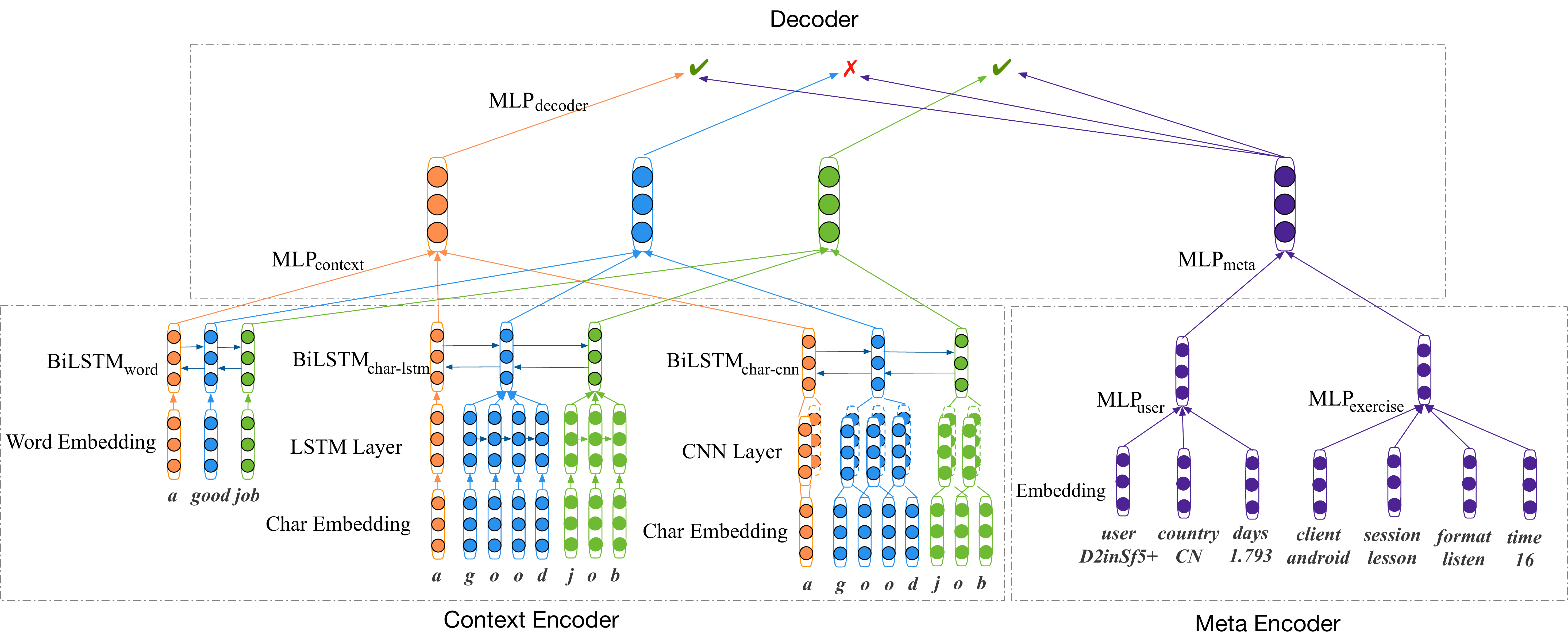}
\caption{ Illustration of our encoder-decoder structure}
\label{Fig:model}
\end{figure*}

Suppose there are $N$ second language-learning datasets $\{D^{1},D^{2},..,D^{N}\}$, and the $k^{th}$ dataset $D^{k}$ is composed of $M^{k}$ exercises $\{e^{k}_{1},e^{k}_{2},...,e^{k}_{M^{k}}\}$,
where $e^{k}_{j}$ is the $j^{th}$ exercise in the $k^{th}$ dataset.

There are two kinds of information in an exercise $e^{k}_{j}$, i.e., the meta information and the language related context information.

The meta information contains two user-related information: (1) user: the unique identifier for each student, e.g., \textit{D2inf5}, (2) country: student's country, e.g., \textit{CN}, and the following five exercise-related information: (1) days: the number of days since the student started learning this language, e.g., \textit{1.793}, (2) client: the student's device platform, e.g., \textit{android}, (3) session: the session type, e.g., \textit{lesson}, (4) format (or type): exercise type, e.g., \textit{Listen}, (5) time: the amount of time in seconds it took for the student to construct and submit the whole answer, e.g., \textit{16s}. This is shared among all language datasets.

The information of the context in the exercise $e^{k}_{j}$ includes the word sequence, that is  $\{w^{1}_{e^{k}_{j}},w^{2}_{e^{k}_{j}},...,w^{l}_{e^{k}_{j}}\}$, and word's linguistic sequences, such as $\{p^{1}_{e^{k}_{j}},p^{2}_{e^{k}_{j}},...,p^{l}_{e^{k}_{j}}\}$, which is the POS-tagging of each word. This is unique to each language-learning dataset.

At last, $e^{k}_{j}$ has a word level label sequence ${\{y^{1}_{e^{k}_{j}},y^{2}_{e^{k}_{j}},...,y^{l}_{e^{k}_{j}}\}}$, where $y_{e^{k}_{j}} \in \{0,1\}$. $y_{e^{k}_{j}}=0$ means this word is answered correctly, and $y_{e^{k}_{j}}=1$ means the opposite. 

Our task is to build a model based on users' exercises, and further to predict word-level label sequence of future exercises.

\subsection{Encoder and Decoder Structure}
Our model is an encoder-decoder structure with two encoders, i.e., a meta encoder, a context encoder, and a decoder. We use the meta encoder to learn the non-linear relationship between meta information, use the context encoder to learn the representation of a sequence of words and use the decoder to generate the final prediction of each word.
The overall structure of the proposed model is shown in Fig.~\ref{Fig:model}.

\textbf{Meta Encoder}:
The meta encoder is a multi-layer perceptron (MLP) based neural network. 
This encoder takes the metadata as inputs.
First, these inputs are converted into high-dimensional representations by the embedding layers, which are randomly initialized and will map each input into a 150-dimensional vector. After the embedding step,
we separately concatenate the user-related embeddings
and the exercise-related embeddings,
and send them into $MLP_{user}$ and $MLP_{exercise}$ to get the representation of user-related meta information $r^{user}$ and the representation of exercise-related meta information $r^{exercise}$, respectively.
Finally, we concatenate  $r^{user}$ and $r^{exercise}$,
and send the concatenated result to $MLP_{meta}$ to obtain the representation of whole meta information $r^{meta}$.
The meta encoder can be formulated as
\begin{equation}
  \begin{aligned}
  & s = [ x^{user}, x^{countries}, x^{days}] \\
  & r^{user} = MLP_{user}(s) \\
  & t = [ x^{format}, x^{session}, x^{client}, x^{time}] \\
  & r^{exercise} = MLP_{exercise}(t) \\
  & r^{meta} = MLP_{meta}([r^{user},r^{exercise}])
  \end{aligned}
\end{equation}
where for the sake of simplicity, 
the variables are omitted from the subscript ${e^{k}_{j}}$,
and $x^{(\cdot)}$ is the embedded representation of each meta information.

\textbf{Context Encoder}:
The context encoder consists of three sub-encoders, i.e., 
a word level context encoder, a char level Long Short Term Memory (LSTM) context encoder, and a char level Convolutional Neural Network (CNN) context encoder.
The word level encoder can capture better semantics and longer dependency than the character level encoders \cite{xu2018cluf}.
By modeling the character sequence, we can partially avoid the out-of-vocabulary (OOV) problem \cite{luong2014addressing}.
Furthermore, we only use the word sequence in the datasets without using any of the provided linguistic information here.
The previous work \cite{rich2018modeling} has pointed out that the linguistic information given by the datasets has mistakes. 
So, through two character level encoders,
we can learn certain word information and linguistic rules.

Given the word sequence 
$\{w^{1}_{e^{k}_{j}},w^{2}_{e^{k}_{j}},...,w^{l}_{e^{k}_{j}}\}$, the 
word level context encoder is computed as
\begin{equation}
  \begin{aligned}
  & x_{t} = Embedding^{word}(w_{t}) \\
  & (g_{1},g_{2}..,g_{l}) = BiLSTM_{word}(x_{1},x_{2},..,x_{l}) \\
  \end{aligned}
\end{equation}
where $w_{t}$ is the $t^{th}$ word in the sequence, and $Embedding^{word}$ is the word embedding. Here, we use the pre-trained ELMo \cite{Peters:2018} as the look-up table. $g_{t}$ is the concatenated result of the last layer's $t^{th}$ hidden state of the forward and the backward cells of $BiLSTM_{word}$. It is also the output of the word level context encoder.

The char level LSTM context encoder is computed according to the sequence characters of word $w_{t} = \{c_{1}, c_{2}, ..., c_{M}\}$. This can be formulated as
\begin{equation}
  \begin{aligned}
        &m_{i} = Embedding^{char}(c_{i}) \\
        &\hat{h}_{w_{t}} = LSTM(m_{1},m_{2},..,m_{l}) \\
        & (\hat{g}_{1},..,\hat{g}_{l}) = BiLSTM_{char-lstm}(\hat{h}_{w_{1}},..,\hat{h}_{w_{l}}) 
      \end{aligned}
\end{equation}
where $\hat{h}$ is the last hidden state of the last layer of $LSTM$.
$\hat{g_{t}}$ is the concatenated result of the last layer's $t^{th}$ hidden state of the forward and the backward cells of $BiLSTM_{char-lstm}$. It is also the output of the char level LSTM context encoder.

The char level CNN context encoder can be similarly formulated as
\begin{equation}
  \begin{aligned}
    &\tilde{h}_{w_{t}} = CNN(m_{1},m_{2},..,m_{l}) \\
    &(\tilde{g}_{1},...,\tilde{g}_{l}) = BiLSTM_{char-cnn}(\tilde{h}_{w_{1}},...,\tilde{h}_{w_{l}}) 
  \end{aligned}
\end{equation}
where $\tilde{h}$ is the result of CNN encoder.
$\tilde{g_{t}}$ is the concatenated result of the last layer's $t^{th}$ hidden state of the forward and the backward cells of $BiLSTM_{char-cnn}$. It is also the output of the char level CNN context encoder.

The final output of the context encoder is generated by a single-layer MLP, and the concatenation of $g_{t}$, $\hat{g}_{t}$ and $\tilde{g}_{t}$ is fed as the input. The process is formulated as
\begin{equation}
  \begin{aligned}
  & r^{context}_{t} = MLP_{context}([g_{t},\hat{g}_{t},\tilde{g}_{t}])
  \end{aligned}
\end{equation}
where $r^{context}_{t}$ is the final context representation of the word $w_{t}$.

\textbf{Decoder}:
The decoder takes the output of meta encoder $r^{meta}$ and 
the output of context encoder $r^{context}_{t}$ as inputs,
the prediction of word $w_{t}$ is computed with a MLP. It is formulated as
\begin{equation}
  \begin{aligned}
  & p_{t} = MLP_{decoder}([r^{context}_{t},r^{meta}])
  \end{aligned}
\end{equation}
where the activation function of $MLP_{decoder}$ is sigmoid function.

\subsection{Multi-Task Learning}

Suppose there are $N$ languages, and each has a corresponding dataset, i.e., $\{D_{1}, D_{2},...,D_{N}\}$. Since our task is to predict the exercise accuracy of language learners on each language, we can regard these predictions as different tasks. Therefore, there are $N$ tasks.

We defined the cross-entropy loss for each task, which encourages the correct predictions and punishes the incorrect ones. Specifically, for the $k^{th}$ task, we have
\begin{equation}
  \begin{aligned}
Loss_{D_{k}} &=-\frac{1}{N}\sum_{t=1}^{N}(\alpha y_{t} \cdot log(p_{t}) \\
&+ (1-\alpha)(1-y_{t})\cdot log(1-p_{t})) \\
  \end{aligned}
\end{equation}
where $\alpha$ is the hyper parameter to balance the negative and positive samples.

In multi-task learning, the parameters in meta encoder and decoder are shared,
and each task only has its own parameters of the context encoder part, so the whole model has only one meta encoder, one decoder and $N$ context encoders. In this way, the common patterns extracted from all language datasets can be utilized simultaneously by the shared meta encoder and decoder.

In the training process, one mini batch contains data of $N$ datasets and they will all be sent to the same meta encoder and decoder, but will be sent to their corresponding context encoder according to their language type.
Thus, the final loss with $N$ tasks is calculated as
\begin{equation}
  \begin{aligned}
Loss_{final} = \sum_{k=1}^{N}Loss_{D_{k}}
  \end{aligned}
\end{equation}
Finally, we use Adam algorithm \cite{kingma2014adam} to train the model.

\section{Experiments}
\subsection{Datasets and Settings}
\begin{table}[t!]
\caption{The statistics of Duolingo SLA modeling dataset }
  \centering
  \small
  \begin{tabular}{llll}
    \hline
    & \textit{en\_es} & \textit{es\_en} & \textit{fr\_en} \\
    \hline
    \#Exercises (Train) &  824,012 & 731,896 &  326,792 \\
    \#Exercises (Dev) &  115,770 & 96,003 &  43,610 \\
    \#Exercises (Test) &  114,586 & 93,145 &  41,753 \\
    \#Unique words &  2,226 &  2,915 & 2,178 \\
    \#Unique users & 2,593 &  2,643 & 1,213 \\
    \#words / exercise &  3.18 &  2.7 & 2.84 \\
    \%OOV radio (Test) & 4.5\% & 10.0\% & 5.9\% \\
    \%Correct radio &  87\% &  86\% & 84\% \\
    \%Incorrect radio &  13\% & 14\% & 16\% \\
    \hline
  \end{tabular}
  
  \label{Tab:data}
  \end{table}
We conduct experiments on Duolingo SLA modeling shared datasets, which have three datasets and are collected from English students who can speak Spanish
(\textit{en\_es}), Spanish students who can speak
English (\textit{es\_en}), and French students who
can speak English (\textit{fr\_en}) \cite{settles2018second}. 
Table~\ref{Tab:data} shows basic statistics of each dataset. 

We compare our method with the following state-of-the-art baselines:
\begin{itemize}
  \item \underline{LR} Here, we use the official baseline provided by Duolingo \cite{settles2018second}. It is a simple logistic regression using all the meta information and context information provided by datasets. 
  \item \underline{GBDT} Here, we use NYU's method \cite{rich2018modeling}, which is the best method among all tree ensemble methods. It uses an ensemble of GBDTs with existing features of dataset and manually constructed features based on psychological theories. 
  \item \underline{RNN} Here, we use singsound's method \cite{osika2018second}, which is the best method among all sequence modeling methods. It uses an RNN architecture which has four types of encoders, representing different types of features: token context, linguistic information, user data, and exercise format.
  \item \underline{ours-MTL} It is our encoder-decoder model \textbf{without} multi-task learning. Thus, we will separately train a model for each language-learning dataset.
\end{itemize}

In the experiments, the embedding size is set to 150 and the hidden size is also set to 150. Dropout \cite{srivastava2014dropout} regularization is applied, where the dropout rate is set to 0.5. We use the Adam optimization algorithm with a learning rate of 0.001.
\subsection{Metric}
SLA modeling is actually the word level classification task, so we use area under the ROC curve (AUC) \cite{hanley1982meaning} and $F_{1}$ score \cite{goutte2005probabilistic} as evaluation metric.

\begin{itemize}
\item AUC is calculated as:
\begin{equation}
  \begin{aligned}
AUC = P(s(x_{1}) > s(x_{2}))
  \end{aligned}
\end{equation}
where $P(\cdot)$ is the probability, $s(\cdot)$ is the trained
classifier, $x_{1}$ is the instance randomly extracted from positive samples, and $x_{2}$ is the instance randomly extracted from negative samples.

\item $F_{1}$ is calculated as
\begin{equation}
  \begin{aligned}
F_{1} = 2 \times \frac{precision * recall}{precision + recall}
  \end{aligned}
\end{equation}
where $precision$ and $recall$ are the precision rate and recall rate of the trained model.
\end{itemize}

\subsection{Experiment on Small-scale Datasets}
\begin{figure*}[t!]
  \centering
  \includegraphics[width=1.0\textwidth]{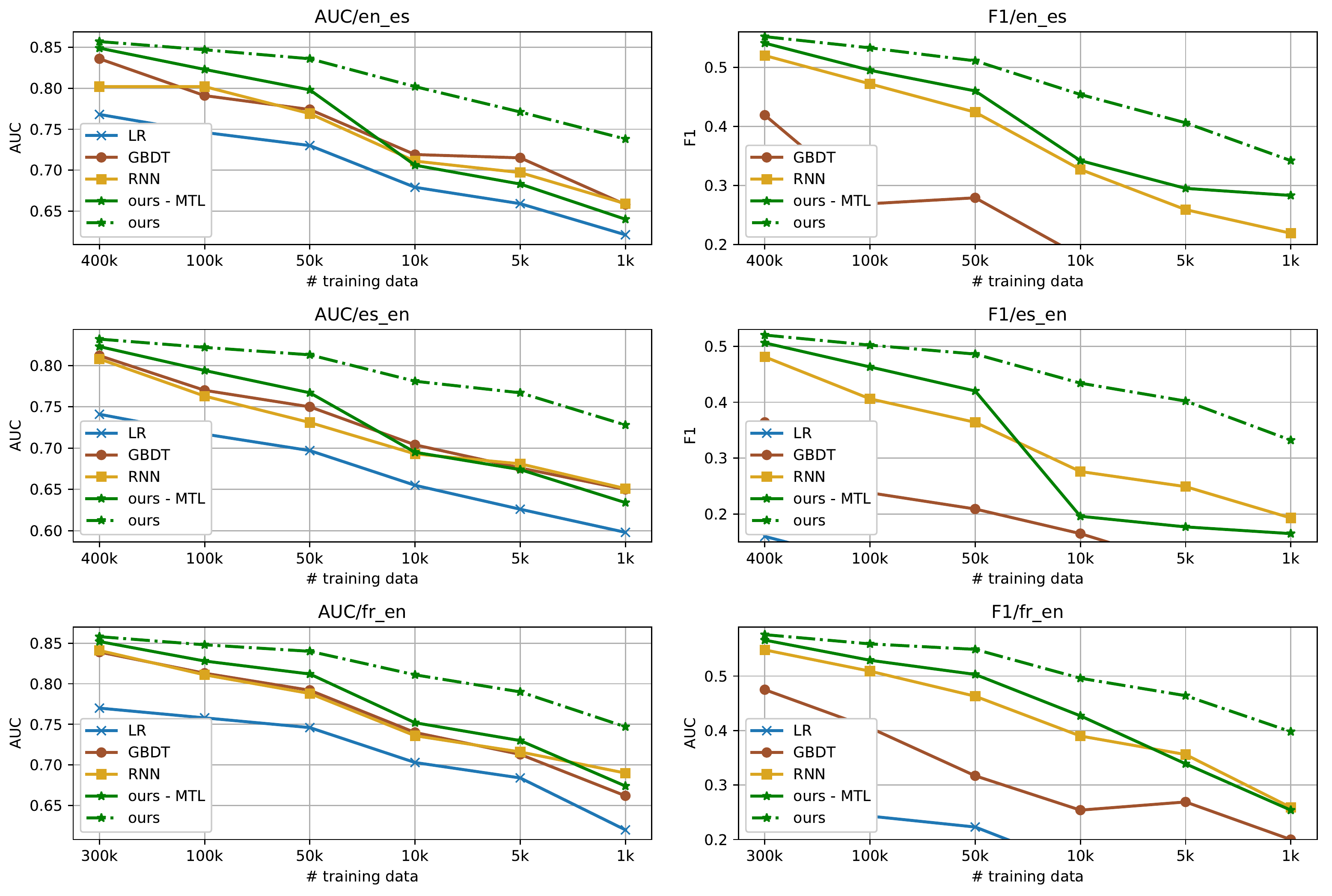}
\caption{Comparison of our method and baselines on training data of different sizes}
\label{Fig:small_data}
\end{figure*}
We first verify the advantages of our method in cases where the training data of the whole language-learning dataset is insufficient. 

Specifically, we gradually decrease the size of training data from 400K ( 300K for \textit{fr\_en} ) to 1K and keep the development set and test set. For all baseline methods, since they only use the single language dataset for training, we hence only reduce the data of corresponding language data. For our multi-task learning method, we reduce the training data of one language dataset and keep the remaining other two datasets unchanged.

The experimental results are shown in Fig.~\ref{Fig:small_data}. 
It can be found that our method outperforms all the state-of-the-art baselines when the training data of a language dataset is insufficient, which is a huge improvement compared with the existing methods.
For example, as shown in \textit{AUC/en\_es} in Fig.~\ref{Fig:small_data}, using only 1K training data, our multi-task learning method 
still could get the AUC score of 0.738, while the AUC score of \underline{ours-MTL} is only 0.640, and existing \underline{RNN}, \underline{GBDT} and \underline{LR} methods are 0.659, 0.658 and 0.650 respectively.
Therefore, the performance of introducing the multi-task learning \textbf{increases by nearly ten percent}. Moreover, to achieve the same performance as our multi-task learning on 1K training data, the methods without multi-task learning require more than 10K training data, which is \textbf{ten times more than ours}.
Thus, multi-task learning utilizes data from all language-learning datasets simultaneously and effectively alleviate the problem of lacking data in a single language-learning dataset.

At the same time, we notice that \underline{ours-MTL} is slightly worse than the \underline{RNN} and \underline{GBDT} when the amount of training data is very small (1K, 5K, 10K). This is because our model does not utilize the linguistic related features of the dataset, and the deep model will be over-fitting when the amount of training data is insufficient. However, as the training data improves ($>$10K), \underline{ours-MTL} becomes better than the existing \underline{RNN} and \underline{GBDT}. Thus, our encoder-decoder structure is very competitive with existing methods even without multi-task learning.


\subsection{Experiment in the Cold Start Scenario}
\begin{table}[t!]
\caption{The statistics of two users (the following number is the number of words in exercises)}
  \centering
    \begin{tabular}{|c|c|c|c|c|}
  \hline
    User & Dataset & Train & Dev & Test \\
\hline
\multirow{2}{*}{\textit{RWDt7srk}} & \textit{es\_en}  & 361 & 68 & 19 \\ 
\cline{2-5} 
& \textit{fr\_en}  & 519 & 80 & 51 \\
\hline
\multirow{2}{*}{\textit{t6nj6nr/}} & \textit{es\_en}  & 562 & 245 & 274 \\ 
\cline{2-5} 
& \textit{fr\_en}  & 998 & 0 & 0 \\
\hline
     \end{tabular}
  \label{Tab:zero_shot}
\end{table}
\begin{table}[t!]
\caption{Comparison of our method and baselines in the cold start scenario}

\centering
    \begin{tabular}{|c|c|c|c|c|}
  \hline
  \multicolumn{3}{|c|}{Methods} & AUC & $F_{1}$ \\
     \hline
     \multicolumn{3}{|c|}{LR \cite{settles2018second}} & 0.765 & 0.083 \\
     \hline
     \multicolumn{3}{|c|}{GBDT \cite{rich2018modeling}} & 0.751 & 0.187 \\
     \hline
     \multicolumn{3}{|c|}{RNN \cite{osika2018second}} & 0.771 & 0.276 \\
     \hline
     \multicolumn{3}{|c|}{ours-MTL} & 0.770 & 0.210 \\
     \hline
     \multicolumn{3}{|c|}{ours} & \textbf{0.881} & \textbf{0.411} \\
     \hline
     \end{tabular}
  \label{Tab:zero_shot_result}
  \end{table}
Further, we can consider directly predicting a user's answer on a language without any training exercises of this user on this language at all. This is cold start scenario and also the situation that the language-learning platforms must consider.

Specifically, it can be found that user \textit{RWDt7srk} and \textit{t6nj6nr/} are all English speakers and learn both Spanish and French, so they have data both in the dataset \textit{es\_en} and \textit{fr\_en}. The statistics are shown in Table \ref{Tab:zero_shot}.
For baseline methods, we remove the data of these two users on the training set as well as development set of \textit{es\_en}, and then train a model. At last, we use the trained model to directly predict the data of this two users on the \textit{es\_en} test set.
Similarly, we use our multi-task method to do the same experiment, and the training data of these two users is also removed from the \textit{es\_en} data set, but \textit{fr\_en} and \textit{en\_es} are unchanged.

The experimental results are shown in Table \ref{Tab:zero_shot_result}. 
If we do not use multi-task learning to predict the new users directly, the performance will be very poor. Compared with the method without multi-task learning, such as \underline{ours-MTL},  our multi-task learning method increases by \textbf{11\%} on ACU and \textbf{20\%} on $F_{1}$.
Because of the multi-task learning, the user information of these two users has been learned through the \textit{fr\_en} dataset. 
Therefore, although there is no training data of these two users on \textit{es\_en}, we can still obtain good performance with mult-task learning.

\subsection{Experiment in the Non-low-resource Scenario} 
\begin{table*}[t!]
 \caption{Comparison of our method with existing methods on different language datasets}
  \centering
  \begin{tabular}{|c|c|c|c|c|c|c|}
    \hline
    \multirow{2}{*}{Methods} & \multicolumn{2}{|c|}{\textit{en\_es}} &  \multicolumn{2}{|c|}{\textit{es\_en}} &  \multicolumn{2}{|c|}{\textit{fr\_en}}   \\
    \cline{2-7}
     &  AUC & $F_{1}$ &  AUC & $F_{1}$ &  AUC  & $F_{1}$  \\
    \hline
    LR \cite{settles2018second} & 0.774 & 0.190 & 0.746 & 0.175 & 0.771 & 0.281 \\
    \hline
    GBDT\cite{rich2018modeling} & 0.859	& 0.468 & 0.835 & 0.420 & 0.854	& 0.493 \\
    \hline
    RNN \cite{xu2018cluf} & 0.861 & 0.559 & 0.835 & 0.524 & 0.854 & 0.569 \\
    \hline
    GBDT+RNN \cite{osika2018second} & 0.861 &	0.561 & 0.838 & \textbf{0.530}	& 0.857 & 0.573 \\
    \hline
    ours-MTL & 0.863 & \textbf{0.564} & 0.837 & 0.527 & 0.857 & 0.575 \\
    \hline
    ours & \textbf{0.864} & \textbf{0.564} & \textbf{0.839} & \textbf{0.530} & \textbf{0.860} & \textbf{0.579} \\
    \hline
  \end{tabular}
  \label{Tab:Result}
  \end{table*}
The experiments above show that our method has a huge advantage over the existing methods in low-resource scenarios. In this section, we will observe the performance of our method in the non-low-resource scenario.

Specifically, we use all the data on the three language datasets to compare our methods with existing methods. This experiment is exactly 2018 public SLA modeling challenge held by Duolingo.\footnote{http://sharedtask.duolingo.com/2018.html}
Here, we add a new baseline \underline{GBDT+RNN}. This is SanaLabs's method \cite{osika2018second} which combines the prediction of a GBDT and an RNN, and it is also the current best method on the 2018 public SLA modeling challenge.

As shown in Table \ref{Tab:Result}, it can be found that although the improvement is not very big, our method surpasses all existing methods on all three datasets and refreshes the best scores on all three datasets.
Especially for the smallest dataset \textit{fr\_en}, our method obtains the most improvement than \underline{ours-MTL}.
As for the largest dataset \textit{en\_es}, our method also improves the AUC score by 0.003 over the best existing method \underline{GBDT+RNN}.
Therefore, our method also gains improvement slightly in the non-low-resource scenario.
\section{Model Analysis}

\subsection{Component Analysis}
\begin{table*}[t!]
\caption{Comparison of encoder removal}
  \centering
  \small
  \setlength\tabcolsep{2pt}
  \begin{tabular}{|c|c|c|c|c|c|c|}
    \hline
    \multirow{2}{*}{Methods} & \multicolumn{2}{|c|}{\textit{en\_es}} &  \multicolumn{2}{|c|}{\textit{es\_en}} &  \multicolumn{2}{|c|}{\textit{fr\_en}}   \\
    \cline{2-7}
     &  AUC & $F_{1}$ &  AUC & $F_{1}$ &  AUC  & $F_{1}$  \\
    \hline
    ours \textbf{remove}  meta encoder & 0.743 & 0.353 & 0.716 & 0.320 & 0.750 & 0.478 \\
    \hline
    ours \textbf{remove} word level context encoder & 0.862 & 0.559 &  0.838 & 0.526 & 0.858 & 0.575 \\
    ours \textbf{remove} char level LSTM context encoder & 0.863 & 0.563 & 0.838 & 0.526 & 0.860 & 0.579 \\
    ours \textbf{remove} char level CNN context encoder & 0.863 & 0.564 & 0.838 & 0.528 & 0.860 & 0.559 \\
    ours \textbf{remove} char level context encoder all & 0.863 & 0.562 & 0.838 & 0.526 & 0.859 & 0.579 \\
    \hline
    ours  & \textbf{0.864} & \textbf{0.564} & \textbf{0.839} & \textbf{0.530} & \textbf{0.860} & \textbf{0.579} \\
    \hline
     \end{tabular}
  \label{Tab:FeatureRemove}
 \end{table*}  

Our encoder-decoder structure contains two encoders, i.e.,  meta encoder and context encoder, where the context encoder includes three encoders, i.e.,  word level context encoder, char level LSTM context encoder and char level CNN context encoder.
In order to explore the importance of each encoder, we do a component removal analysis experiment.

Specifically, we remove each encoder component, train a model, and record the performance on test set. We also remove both two char level context encoders and do the same experiment.

The experimental results are shown in Table \ref{Tab:FeatureRemove}.
It can be found that the meta information is critical to the final result, much more important than the context encoder. If the meta encoder is removed, the result will be sharply reduced. 
The reason is that: if there is only a context encoder, it is equal to modeling the global word error distribution, completely ignoring the individual's situation, which violates adaptive learning.

For context encoder, word level encoder has a greater impact than char level encoder on the performance of our model.
\subsection{Metadata Analysis}
The analysis above has proven that meta information is important for predicting results. Obviously, different features of meta information have different influence. Therefore, feature removal analysis is made to find important features. Specifically, we remove each meta feature and get the performance of the model without this feature.

\begin{figure}[t!]
  \centering
  \includegraphics[width=0.5\textwidth]{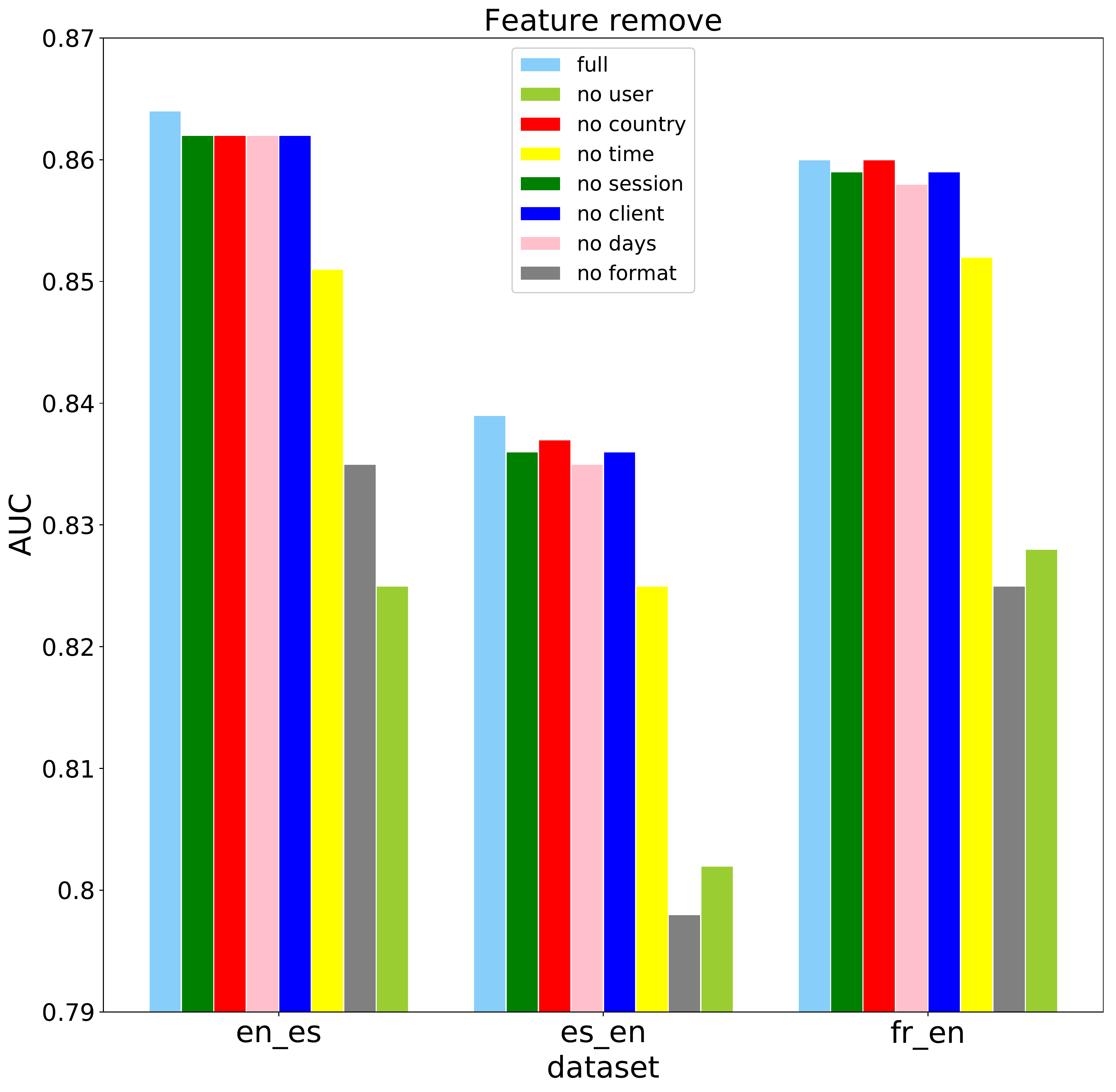}
\caption{Analysis of meta features removal}
\label{Fig:feature remove}
\end{figure}

As shown in Fig.~\ref{Fig:feature remove}, the most important feature is the user (id). Without user (id), the model performance declines rapidly, because user information is the key to building user-adaptive learning.
This also shows that the most common pattern between learning different languages is the students themselves.
Besides, it can be found that learning format and spent time also make significant influences on the model.

\subsection{Visualization}
In this part, we will show what meta encoder has learned from three datasets by multi-task learning.
\begin{figure}[t!]
  \centering
  \includegraphics[width=0.5\textwidth]{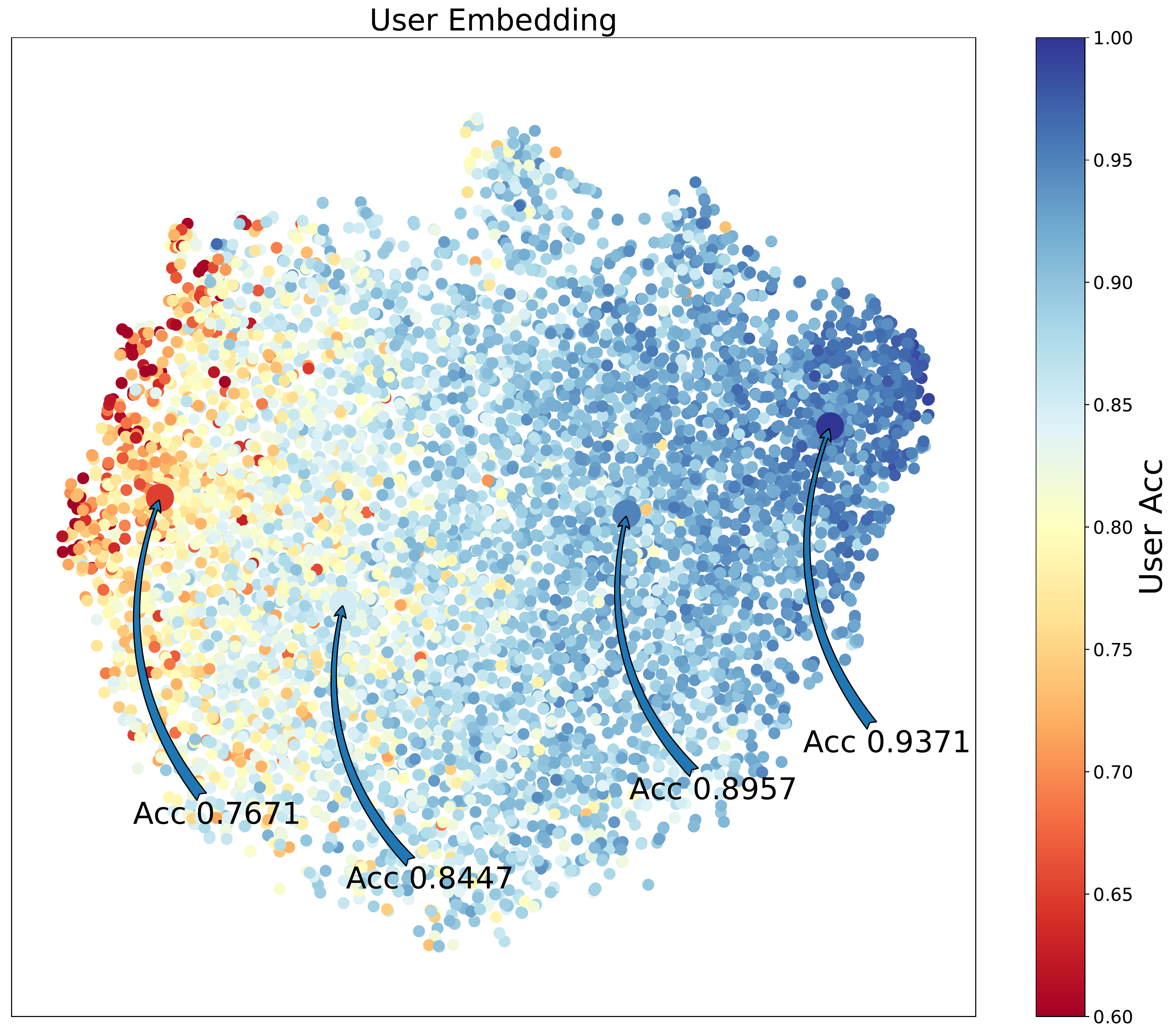}
\caption{User embedding cluster}
\label{Fig:user}
\end{figure}

We cluster the user embedding with k-means algorithm ($k=4$), and calculate the average accuracy of each user and the overall average accuracy of each cluster.
Embeddings are processed by t-SNE \cite{maaten2008visualizing} for visualization, as shown in Fig.~\ref{Fig:user}, every point represents a user and its color represents the average accuracy of this user. Red means low accuracy and blue means high. The four large points indicate the center of clustering, and the value pointing to the point is the overall average accuracy of the corresponding cluster.
It can be found that students with good grades and students with poor grades can be distinguished very well according to their user embeddings, so the user embedding trained by our model contains rich information for the final prediction.


\section{Conclusion}
In this paper, we have proposed a novel multi-task learning method for SLA modeling. As far as we know, this is the first work applying multi-task neural network to SLA modeling and study the common patterns among different languages. Extensive experiments show that our method performs much better than the state-of-the-art baselines in low-resource scenarios, and it also obtains improvement slightly in the non-low-resource scenario.

\bibliography{reference}
\end{document}